\documentclass[conference]{IEEEtran}

\IEEEoverridecommandlockouts
\usepackage{cite}
\usepackage{float}
\usepackage{amsmath,amssymb,amsfonts}
\usepackage{algorithmic}
\usepackage{graphicx}
\usepackage{textcomp}
\usepackage{xcolor}
\usepackage{multirow}

\newtheorem{definition}{Definition}

\def\BibTeX{{\rm B\kern-.05em{\sc i\kern-.025em b}\kern-.08em
    T\kern-.1667em\lower.7ex\hbox{E}\kern-.125emX}}
\begin{document}

\title{Deep Probability Segmentation: Are segmentation models probability estimators? \\ 
}

\author{\IEEEauthorblockN{Simone Fassio}
\IEEEauthorblockA{\textit{Politecnico di Torino} \\
Torino, Italy \\
simone.fassio@studenti.polito.it}
\and
\IEEEauthorblockN{Simone Monaco, Daniele Apiletti}
\IEEEauthorblockA{\textit{Deparment of Control and Computing engineering, Politecnico di Torino} \\
Torino, Italy \\
\{name\}.\{surname\}@polito.it}
}

\maketitle

\begin{abstract}
Deep learning has revolutionized various fields by enabling highly accurate predictions and estimates. One important application is probabilistic prediction, where models estimate the probability of events rather than deterministic outcomes. This approach is particularly relevant and, therefore, still unexplored for segmentation tasks where each pixel in an image needs to be classified. Conventional models often overlook the probabilistic nature of labels, but accurate uncertainty estimation is crucial for improving the reliability and applicability of models.

In this study, we applied Calibrated Probability Estimation (CaPE) to segmentation tasks to evaluate its impact on model calibration. Our results indicate that while CaPE improves calibration, its effect is less pronounced compared to classification tasks, suggesting that segmentation models can inherently provide better probability estimates. We also investigated the influence of dataset size and bin optimization on the effectiveness of calibration. Our results emphasize the expressive power of segmentation models as probability estimators and incorporate probabilistic reasoning, which is crucial for applications requiring precise uncertainty quantification.
\end{abstract}

\begin{IEEEkeywords}
deep probability estimation, image segmentation, convolutional neural networks
\end{IEEEkeywords}

\section{Introduction}



Machine learning, particularly deep learning, has transformed various fields by enabling models to achieve exceptional accuracy in predictions and estimates. This success is due to its ability to extract latent patterns from large data sets and generalize them to new cases. A widely used application of deep learning is supervised learning, where models learn to map the input data to a ground truth that represents the target outcome. Traditionally, this mapping is viewed as deterministic without considering the probabilistic nature that these labels may inherently have. However, this view changes when models answer questions such as \textit{will it rain heavily?} or \textit{will a forest fire affect a region?} where the answer is inherently probabilistic. Models developed for probabilistic predictions need to be trained on observed outcomes, which are accessible parts of the latent probability distribution. This paradigm shift, which involves aleatory uncertainty quantification within model prediction, has received only moderate attention, but traditional models generally yield inconsistent results~\cite{liu2021deep}.

Besides this, an emerging and increasingly popular variant of classification tasks is segmentation, where the goal is to assign a class to each pixel in an input image. The probabilistic interpretation discussed above can of course also be applied to segmentation, but with the added complexity that spatial relationships between the pixels must be taken into account. Neighboring pixels often share contextual information that influences their labeling. However, as the image size increases, so does the uncertainty associated with these labels. This is due to the potential aleatory uncertainty arising from both the underlying probability distribution and annotation errors.

Considering this probabilistic view during segmentation is crucial for improving the reliability and applicability of the resulting models in practical scenarios. Accurate uncertainty estimation improves the robustness of segmentation results and provides valuable insights into the confidence levels of predictions. This capability is particularly important in areas where accurate segmentation supports decision making, risk assessment, and scientific research. Therefore, advanced methods that integrate probabilistic reasoning into segmentation models hold great promise for improving their effectiveness and versatility in various applications. Based on previous studies using neural networks as probability estimators~\cite{liu2021deep}, we aim to extend this idea to segmentation tasks. This will allow us to investigate the impact of pixel correlations and assess how the latest models in this field can perform this type of task.

Our study applied Calibrated Probability Estimation (CaPE) to segmentation tasks to assess its impact on improving model calibration. Although CaPE led to improvements in calibration, its effectiveness appeared somewhat subdued compared to its performance on classification tasks. This observation suggests that many segmentation datasets inherently provide reasonably accurate probability estimates without additional calibration interventions. We also investigated possible correlations between the target event's probability distribution and the training dataset's size. Furthermore, we performed detailed investigations on optimizing the number of bins, a critical hyperparameter within the CaPE method.

By delving deeper into these aspects, we can not only evaluate the direct benefits of probabilistic modelling in segmentation, but also understand the nuanced factors influencing the quality of probability estimates in this context. This exploration is crucial for improving the reliability and applicability of segmentation models in various real-world scenarios, where accurate uncertainty quantification can significantly improve decision making and domain-specific applications.

The paper is structured as follows: Section~\ref{sec:relworks} gives an overview of related work on probabilistic estimation in neural networks and its application to segmentation tasks. Section~\ref{sec:methods} outlines our methodology, including the implementation of Calibrated Probability Estimation (CaPE) in segmentation, details of the validation dataset and the evaluation metrics used. Section~\ref{sec:results} presents and discusses the results. Section~\ref{sec:conclusions} concludes with a summary of our results and suggestions for future research directions.

\section{Related works}
\label{sec:relworks}

Deep neural networks, especially when applied to classification tasks, often produce probability estimates reflecting the model's confidence. However, these estimates can be inaccurate, leading to poorly calibrated predictions. To overcome this challenge, several methods have been proposed to improve calibration, including post-processing techniques and ensembling.

Post-processing techniques aim to refine output probabilities to improve the calibration of unseen data. Notably, Guo et al.~\cite{guo2017calibration} introduced temperature scaling, incorporating a temperature parameter into the softmax function. This parameter is optimized to minimize negative log-likelihood loss, thereby adjusting the model's confidence levels to improve calibration. Another approach involves training a recalibration model based on the outputs of an uncalibrated model~\cite{kuleshov2018accurate}. These methods have demonstrated effectiveness in enhancing the calibration of deep learning algorithms in a model-agnostic fashion, without altering their fundamental architectures.

Conversely, ensembling techniques combine multiple models to enhance generalization and alleviate miscalibration. For instance, Zhang et al.~\cite{zhang2020mix} proposed Mix-n-Match, which employs a single model and combines predictions using multiple temperature scaling transformations.

Recently, Liu et al. \cite{liu2021deep} introduced Calibrated Probability Estimation (CaPE), a training technique aimed at improving neural networks' ability to generate reliable probability estimates. The method ensures that the output probabilities match the empirical probabilities derived from the data, thereby calibrating the model results when confronted with aleatory uncertainty in the data. While this method has proven successful on various synthetic and real-world classification tasks, it has not yet been applied to tasks with non-independently identically distributed (\textit{i.i.d.}) data, such as image segmentation problems or predictions on graph structures. This work aims to extend this analysis in this direction and show how traditional models behave when confronted with aleatory uncertainty in the data and how calibration affects them.

Deep learning segmentation models have been used in various fields that are inherently fraught with uncertainty. Among these, medical imaging is a prominent application area where researchers have grappled with the challenge of overcoming uncertainty in label assignment. For example, Kohl et al. \cite{kohl2018probabilistic} introduced a probabilistic U-Net, which provides segmentation results along with the uncertainty associated with these predictions. This capability is crucial in medical diagnostics, as understanding the confidence level of detection significantly influences clinical decisions. In addition, Krygier et al. \cite{krygier2021quantifying} investigated the impact of labeling uncertainty in 3D medical images and proposed a method to quantify this uncertainty through Bayesian convolutional neural networks and Monte Carlo dropout. Their results illustrate the complexity of uncertainty distributions in simulations that challenge conventional segmentation boundaries.

In another domain, Hu et al. \cite{hu2023deep} explored the use of deep learning architectures for post-processing deterministic numerical weather predictions of precipitation, focusing on improving forecast accuracy and deriving uncertainty estimates. Their segmentation approach, leveraging distributional parameters of a censored shifted gamma distribution, addresses the uncertainty arising from the numerical simulation inputs. However, the applicability of this method is highly task-dependent.

In addition, Monaco et al. \cite{monaco2021attention} investigated segmentation models for assessing the severity of wildfires, emphasizing the challenge posed by unreliable ground truth labels. Their work emphasizes the importance of understanding the uncertainty of the input data to predict the severity of wildfires accurately.

To date, the capabilities of these segmentation models as deep probability estimators have not been systematically investigated.

\section{Material and methods}
\label{sec:methods}

\subsection{Case studies}
In this study, we use two different datasets to evaluate the performance of our proposed model in different domains. The first dataset is the German Weather Service (GWS) dataset, which focuses on weather forecasting and predicts the probability of exceeding a certain weather threshold at each location. The second dataset was developed specifically for Burned Area Detection (BAD) and estimating the extent of damage. It includes the segmentation of satellite images and aims to determine the probability that the area represented by each pixel is burnt with a predefined severity.

The \textbf{GWS} dataset comprises quality-controlled rainfall-depth composites gathered from 17 operational Doppler radars, providing a comprehensive perspective on precipitation patterns. Our analysis uses 30-minute precipitation data, comprising three precipitation maps from the past 30 minutes, to predict the probability of precipitation exceeding various thresholds in each pixel one hour after the most recent measurement. Given the size of the dataset, we decided to use only part of it, as using the entire dataset would not provide additional valuable information. Therefore, we reduced the number of events to 3000, which we selected from the 15,000 events with the highest average precipitation, which is in line with our task. Furthermore, from the original resolution of 900x900, we only considered the central part of 512x384, which was then scaled to 128x96. This procedure reduced the demand for computational resources, but we observed that it did not significantly impact the overall result.

The \textbf{BAD} dataset contains 73 satellite images of various forests damaged by wildfires across Europe with a resolution of up to 10 meters per pixel. The Sentinel-2 L2A satellite mission collected the data, while the target labels were generated from Copernicus Emergency Management Service annotations, with five severity levels from undamaged to completely destroyed~\cite{colomba2022dataset,monaco2021attention}. The images we used were taken within one month of the wildfire event, ensuring minimal cloud cover. The original images of the dataset are divided into tiles of 128x96 pixels. To remain consistent with the previous dataset, we select only the tiles that contain at least one burned pixel, resulting in 2,691 valid tiles.

Table~\ref{tab:thresholds_probabilities} describes the thresholds that were considered for each dataset and their respective probabilities of being exceeded.

\begin{table}[htbp]
\caption{Thresholds and Probabilities for Each Dataset}
\begin{center}
\begin{tabular}{|c|c|c|}
\hline
\textbf{Dataset} & \textbf{Threshold} & \textbf{Probability (\%)} \\
\hline
\multirow{5}{*}{GWS} & 0.02  & 30  \\
                     & 0.125 & 14  \\
                     & 0.250 & 7   \\
                     & 0.5   & 3.2 \\
                     & 0.75  & 1.1 \\
\hline
\multirow{3}{*}{BAD} & Severity Level 1 & 46 \\
                     & Severity Level 3 & 27 \\
                     & Severity Level 4 & 14 \\
\hline
\end{tabular}
\label{tab:thresholds_probabilities}
\end{center}
\end{table}

\subsection{Problem formulation}

The probability estimation task involves evaluating the likelihood of an event of interest based on the available data. In our context, we consider a dataset comprising $n$ example images $x_i$ of size $D \times W$ along with their corresponding segmentation maps $y_i$ of the same dimensions. Each pixel in $y_i$ is binary, with a 0 or 1 value indicating whether the event occurred at the corresponding position in $x_i$.

To illustrate, consider the example of rainfall estimation beyond a defined threshold in climate forecasting. Here $x_i$ represents meaningful available quantities from a previous time step, while $y_i$ is the binary map stating if the rain quantity exceeded the threshold in each pixel in the grid. The data inherently contains uncertainty. Each pixel's value may depend on previous time steps or unavailable meteorological information. We can consider each pixel $y_i$ to be 1 with a certain probability $p_i$ associated with neighbouring pixels from $x_i$. This is because the input data provides important information about the event's occurrence. Then, a probability-estimation model will be trained to produce the best estimation $\hat p$ of $p$ on new images $x$. It is worth noting that this is not a traditional classification task, as the goal is to predict the probability of the outcome, which inherently involves aleatory uncertainty.

To measure the effectiveness of our model, it is useful to introduce the concept of calibration, which refers to the correspondence between the predicted probabilities and the actual results. We need this concept if we rely on the observed results since the real probability $p$ is unavailable. Then, a probability-estimation model $f_\theta$ parameterized with weights $\theta$, we define as follows:
\begin{definition}
  The parametric model $f_\theta$ is \textit{well-calibrated} if exists a small interval $I(q)$ such that
  \begin{equation}
      \mathcal{P}(y=1|f_\theta(x) \in I(q))=q,\;\; \forall 0\leq q\leq 1,
  \end{equation}
\end{definition}

where $y$ is the observation associated with the input $x$ and $f_\theta(x)$ is the probability predicted by model.

\subsection{Evaluation metrics}
Expected Calibration Error (ECE)~\cite{guo2017calibration} is a standard calibration assessment. To evaluate the probabilities predicted by the model, we compare them with empirically observed probabilities. To do this, we group all pixels $x_i$ where the model's output matches a certain value and calculate the proportion of these pixels with a result of 1. If this proportion is very similar to the probability predicted by the model, the model is considered to be well-calibrated. In our context, we have repeated these considerations for each pixel of the input data $x_i$ and the corresponding results $y_i$. 
Once we gather the outcomes for all pixels $y_i$, we can partition the model's predicted probabilities, represented by $f_\theta(x_i)$ , into $B$ bins: $I_1, I_2, \ldots, I_B$. This partitioning is based on the predicted probability values assigned to each pixel.
Let $Q_1, \ldots, Q_{B-1}$ be the $B$-quantiles of the set $\{f_\theta(x_1), \ldots, f_\theta(x_N)\}$, we have
\begin{equation}
    I_b := [Q_{b-1}, Q_b] \cap \{f_\theta(x_i)\}_{i=1}^N \quad \text{(setting } Q_0 = 0).
\end{equation}

For each bin $I_b^i$, we compute the mean of empirical and predicted probabilities, defined as

\begin{equation}
\label{eq:pemp}
p_\text{emp}^{(b)} = \mathbb{E}\left( y \mid f_\theta(x) \in I_b \right) = \frac{1}{|I_b|} \sum_{i \in \text{Index}\left(I_b\right)} y_i
\end{equation}
\begin{equation}
q^{(b)} = \frac{1}{|I_b|} \sum_{i \in \text{Index}\left(I_b\right)} f_\theta\left(x_i\right)
\end{equation}
where Index($I_b$) = \(\{ i \mid f_\theta(x_i) \in I_b \}\)

\begin{equation}
    \text{ECE} = \frac{1}{B} \sum_{b=1}^B \left|p_\text{emp}^{(b)} - q^{(b)}\right|
\end{equation}

Another important metric for quantifying uncertainty is the Brier score, which measures the mean squared difference between the predicted probabilities and the actual outcomes and provides a single score that summarizes the accuracy of probabilistic predictions. 
\begin{equation}
    \text{Brier} = \frac{1}{N} \sum_{i=1}^{N} (f_\theta(x_i) - y_i)^2
\end{equation}
where $N$ is the number of predictions, $f_\theta(x_i)$ is the predicted probability for the $i$-th pixel, and $y_i$ is the actual binary outcome.
A lower Brier score indicates better performance because it means that the predicted probabilities are closer to the actual results. The Brier score ranges from 0, a perfect probabilistic prediction, to 1, the worst possible deviation from the target probability.

Finally, the Kullback-Leibler (KL) divergence is a widely used evaluation metric in machine learning and information theory, especially for comparing probability distributions. 
It measures the difference between two probability distributions and indicates how much a distribution deviates from a reference distribution.
Mathematically, the KL divergence between a distribution $Q$ (the estimated distribution) and a distribution $P$ (the true distribution) is defined as follows:
\begin{equation}
    D_{\text{KL}}(P \parallel Q) = \sum_{i} P(i) \log \left( \frac{P(i)}{Q(i)} \right)
\end{equation}
In this formula, $P(i)$ represents the true probability of the $i$-th event, and $Q(i)$ represents the predicted probability of the same event. KL divergence is a valuable tool for evaluating the performance of probabilistic models. Quantifying the divergence between the predicted and true distributions clearly measures how well a model captures the underlying data distribution.

\subsection{CaPE regularization}

Based on the definition from Liu et al.~\cite{liu2021deep}, we introduce CaPE regularization to a segmentation pipeline, to evaluate whether it can improve the model performance.
This method is designed to improve the calibration of probabilistic models by minimizing a calibration loss alongside the traditional discrimination loss $\mathcal{L}_D$, generally Binary Cross Entropy (BCE). CaPE strategy initially trains the model on $\mathcal{L}_D$ with early stopping to prevent overfitting. It then employs a weighted sum approach, combining $\mathcal{L}_D$ with an extra term $\mathcal{L}_C$ enforcing model calibration. This second term measures the cross-entropy between the model’s predicted probabilities and the empirical probabilities conditioned on the model output.
The discrimination loss compares the predicted probability for each area with the actual label, penalizing the predictions based on the difference between the inferred probability and the true label. In contrast, the calibration loss compares the predicted probability distribution to the empirical one. This dual optimization aims to enhance both the discrimination ability and the calibration of the model, ensuring that the predicted probabilities align closely with the true likelihood of outcomes. The two loss functions are formalized as follows:

\begin{equation}
    \begin{split}
        \mathcal{L}_{\text{D}} &= - \sum_{i=1}^N \left[y_i \log(f_\theta(x_i)) + (1 - y_i) \log(1 - f_\theta(x_i))\right]\\
        \mathcal{L}_{\text{C}} &= - \sum_{i=1}^{N} \left[ p_{\text{emp}}^i \log \left( f_\theta(x_i) \right) + \left( 1 - p_{\text{emp}}^i \right) \log \left( 1 - f_\theta(x_i) \right) \right]
    \end{split}
    \label{eq_losses}
\end{equation}
where $p^{i}_{\text{emp}}$ is an estimate of the conditional probability \(P[y = 1 \mid f_\theta(x) \in I(f_\theta(x_i))]\) and $I(f_\theta(x_i))$ is a small interval centered at $f(x_i)$. We consider the \textit{bin} version of CaPE method for estimating $p_{\text{emp}}^i$, which involves dividing the training set into bins.

We identify the bin $b_i$ that contains $f_\theta(x)$ and assign $p_{\text{emp}}^i$ as $p_{\text{emp}}^{(b_i)}$ in Equation~\ref{eq:pemp}. This method can be efficiently executed by arranging the predictions $p_i$. The calibration loss necessitates a reasonable estimation of the empirical probabilities $p_{\text{emp}}(i)$, which can be derived from the model after initial training.

\subsection{Experimental design}
To determine the ability of the segmentation model to estimate the probability behind the data, we examine the performance differences between a traditional training pipeline and training the same architecture with the CaPE strategy.

For all experiments, we chose a U-Net~\cite{ronneberger2015u} as the segmentation model because it performs best for similar tasks in limited data domains~\cite{wang2022medical}. Our entire pipeline is not dependent on this choice. We, therefore, also chose this model because it serves as a solid foundation for the segmentation task, and the results obtained with this architecture can potentially be transferred to other related architectures.

Our network has a symmetrical 5-block structure with an encoder and a decoder. The encoding path sequentially reduces the spatial dimensions of the feature maps while increasing the number of channels, starting with 64 channels and doubling with each step until 1024 channels are reached. This process allows the network to capture increasingly abstract and detailed features. The decoder path then upsamples these feature maps back to the original input resolution and reverses the encoder process. Notably, our implementation does not use skip connections, which are typically used in U-Net architectures to combine high-resolution features from the encoder with high-sampled features in the decoder. The final output layer of the network generates a segmentation mask in which the intensity of each pixel indicates the probability that it exceeds the predefined threshold. This structure allows us to effectively analyze and segment the input data based on the learned features.
As a loss function, we use the BCE loss, a common choice for binary segmentation tasks.

The training was performed on a system equipped with 128 GB RAM and an NVIDIA Tesla V100 GPU with 16 GB memory. We trained over 50 epochs with the Adam optimizer and a learning rate of 0.0001. The early stopping strategy is applied to the validation loss, with a patience of 15 epochs and a minimum delta of $0$. To evaluate the model's performance, we used a 9-fold cross-validation approach: 7 folds were used as the training set, 1 fold as the validation set, and 1 fold as the test set. All results are finally aggregated from the test sets of each fold. The code used to perform the experiments is available upon request.

\section{Results and discussions}
\label{sec:results}

\begin{figure}[ht]
    \centering
    \includegraphics[width=\linewidth]{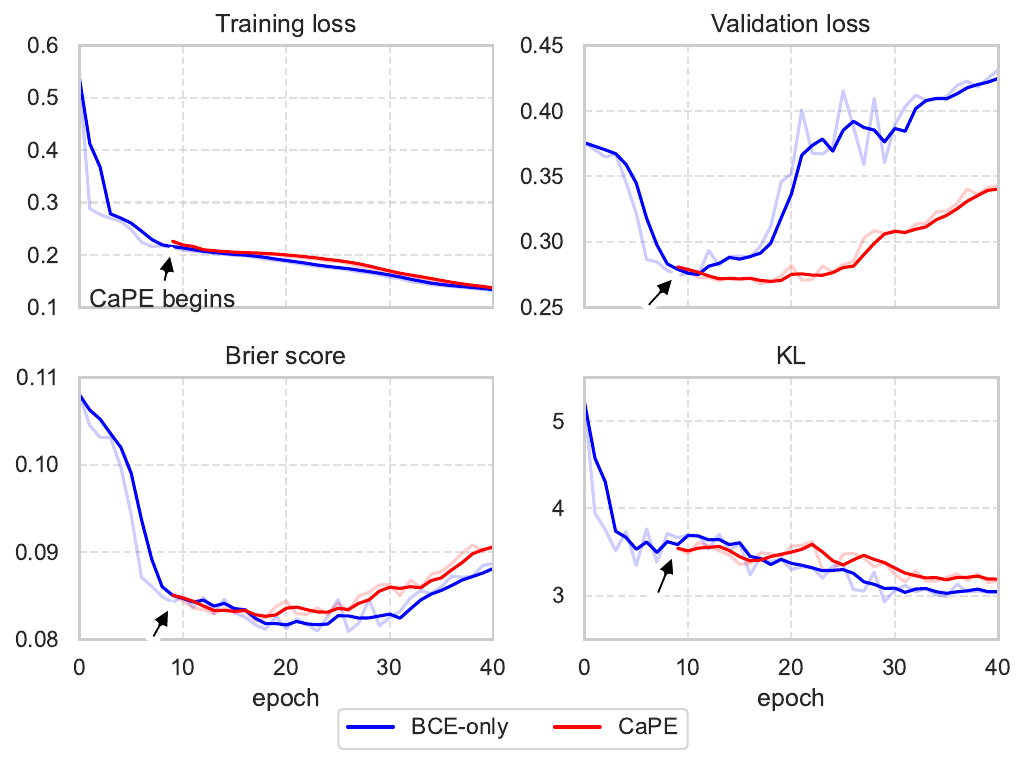}
    \caption{Comparison between the learning curves of BCE minimization and CaPE over 50 epochs, smoothed with a 3-epoch moving average. The numbers from left to right are the training loss, validation loss, Brier score and KL divergence, respectively. The softer lines represent the raw data. The CaPE method starts with the minimum of the validation loss. From this point onwards, the effect compared to BCE training is evident.}
    \label{fig:loss}
\end{figure}

One of the main questions is whether the advantages of the CaPE strategy still apply to the segmentation task. To clarify this point, Figure~\ref{fig:loss} shows the learning curve of the model over the first 50 epochs. The black arrow shows the point at which early stopping detected an overfitting validation loss, triggering the start of CaPE application. The red curves represent a model trained with the combined losses in Equation~\ref{eq_losses}, while the blue curves represent the model trained with $\mathcal{L}_D$ after early stopping. The presented training refers to the GWS dataset with the following parameters: dataset size of 1500, number of bins set to 20, and a threshold of 0.250 mm/h. We present this setting because, according to our experiments, it represents an average case. 
In the remainder of this analysis, we will look in detail at the contribution of calibration on the distribution of dataset labels and other relevant hyperparameters.

We find that both training and validation losses initially decrease. Then, CaPE regularization limits overfitting for about ten epochs after its first application. This indicates that it can stabilize the training and provide slightly better results than the unregularized model. However, this improvement does not reflect a significant reduction in Brier score and KL divergence. This behavior is clearly different from what was observed in non-segmentation tasks, suggesting that segmentation models may be inherently better calibrated than other architectures evaluated in this work.

At this point, we focused our experiments on describing the different contributions of the CaPE method when the threshold, and thus the distribution of the event we want to predict, changes. In particular, we investigated how different threshold settings affect model calibration and uncertainty estimates when the size of the training dataset varies. These two parameters regulate the distribution of positive labels and may provide different contributions to the regularization strategy.

\begin{figure}[ht]
    \centering
    \includegraphics[width=\linewidth]{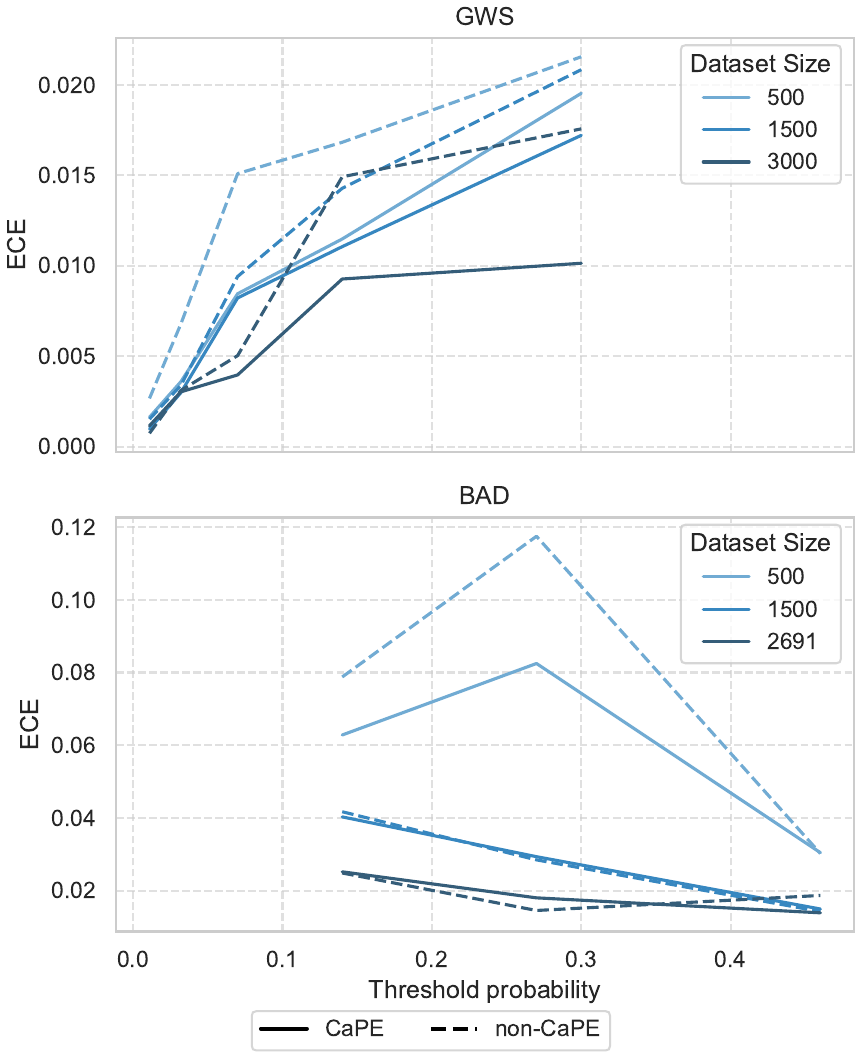}
    \caption{Comparison of the Expected Calibration Error (ECE) for the GWS and BAD datasets across different dataset sizes. The solid lines represent the results after applying the CaPE method, while the dashed lines represent the results with early stopping.}
    \label{fig:ECEvsTH}
\end{figure}

Figure~\ref{fig:ECEvsTH} shows the best value of ECE achieved by the models over the variable threshold probability for both datasets under analysis.
Based on the upper graph, we observe a positive correlation between ECE and the probability of exceeding the threshold in the GWS dataset. This suggests that as the probability of exceeding the weather threshold increases, the calibration error also tends to increase, indicating greater uncertainty in these forecasts. This finding is crucial as it highlights a potential challenge in achieving well-calibrated probabilities under conditions of high uncertainty. In particular, when the model predicts a high probability of extreme weather events, the uncertainty in these predictions becomes larger, leading to higher calibration errors. Remarkably, the discrepancies between CaPE and non-CaPE results increase with the threshold value, especially for curves representing large training sets.

In contrast, the BAD dataset does not reflect the positive correlation between threshold and ECE. In this dataset, there were fewer possible thresholds that were more likely to be exceeded than in the GWS dataset. This discrepancy could be due to the nature of the data and the distribution of events in the BAD dataset. In particular, we can assume that the higher probability of exceeding the thresholds could simplify the prediction task and reduce the calibration error. In addition, the model's difficulty distinguishing between the first and subsequent severity levels could also contribute to this behavior, suggesting that the model may perform better when the event distribution is less complex. It is worth noting that CaPE makes no significant contributions in this dataset, except at the smallest size of the training dataset.

\begin{figure}[ht]
    \centering
    \includegraphics[width=\linewidth]{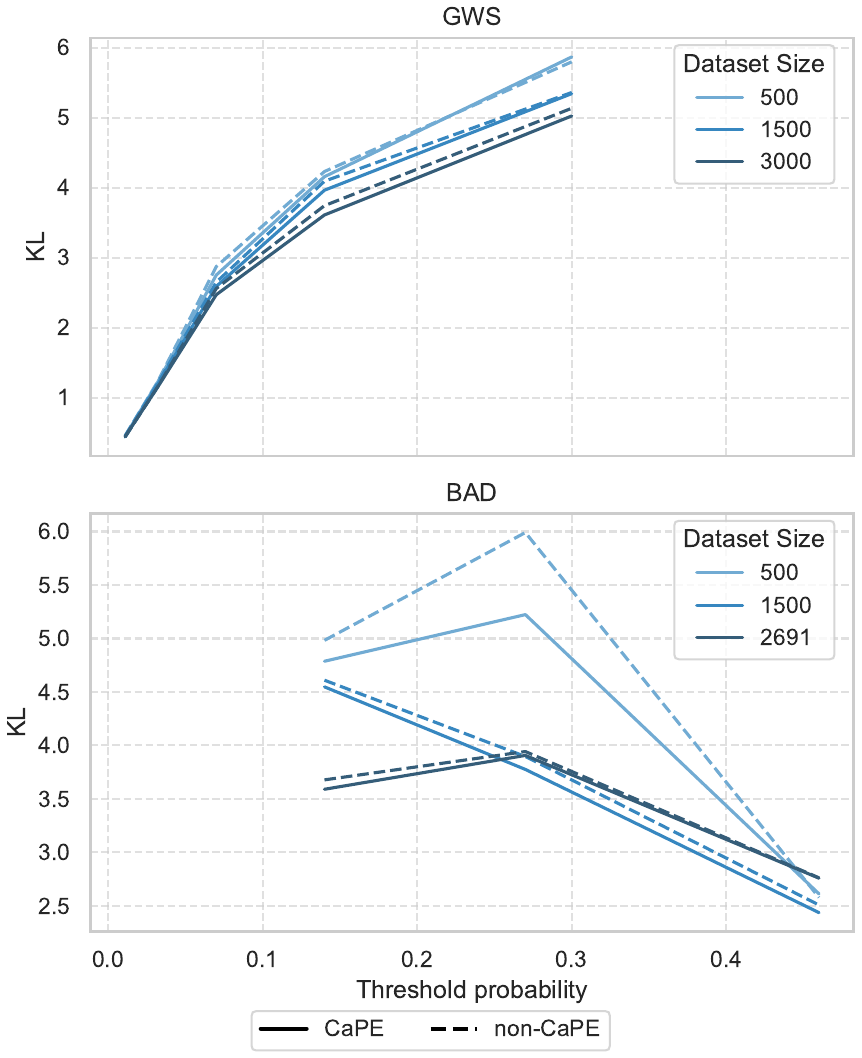}
    \caption{Comparison of KL for the GWS and BAD datasets across different dataset sizes. The solid lines represent the results after applying the CaPE method, while the dashed lines represent the results with early stopping.}
    \label{fig:KLvsTH}
\end{figure}

As for ECE, Figure~\ref{fig:KLvsTH} illustrates the behaviour of KL divergence when the threshold probability varies. We observe a much smaller variation in KL divergence compared to the previous plot, especially for the GWS dataset. Overall, we can conclude that the contribution of CaPE in this metric is negligible in most cases. Furthermore, it is interesting to note that the two metrics are still highly correlated: the trends in the KL divergence graphs are very similar to those in the ECE graphs in Figure ~\ref{fig:ECEvsTH}. This similarity suggests that both metrics capture related aspects of model performance despite their different formulations.
Both Figures ~\ref{fig:ECEvsTH} and ~\ref{fig:KLvsTH} show data for a bin number of 20 for the CaPE method. We found that different numbers of bins in our experiments would lead to similar results overall, apart from minor local differences.

Finally, this analysis also shows no significant correlation between the size of the data set, the threshold, and the number of bins on the effectiveness of the CaPE method. These parameters appear to be highly task-dependent, suggesting that their influence depends on the nature of the task in question. For example, while the number of bins may be critical for one dataset, it may not have the same impact on another dataset with different characteristics. This task dependency emphasizes the importance of tailoring the calibration approach to the specific requirements and characteristics of the dataset.

This observation highlights the effectiveness of the calibration strategy, but also suggests that both regularization against overfitting and calibration of probabilities in segmentation tasks seem to have a smaller effect compared to traditional problems, as shown by Liu et al.~\cite{liu2021deep}. This result suggests the intriguing possibility that segmentation models are inherently better calibrated, possibly due to the non-i.i.d. nature of pixels in image data in such tasks.

\section{Conclusions}
\label{sec:conclusions}

In this study, we investigated the application of the Calibrated Probability Estimation (CaPE) method to segmentation tasks specifically targeting weather forecasting and fire area detection. Our experiments have shown that while CaPE offers some advantages in model calibration, its impact is less pronounced than classification tasks. This result suggests that certain segmentation datasets inherently provide well-calibrated probability estimates.
A key advantage of CaPE was its role in preventing overfitting by acting as a regularization technique. Despite minimal improvements in metrics such as KL and Brier score, CaPE effectively maintained calibration and improved the robustness of the model to overfitting.
Further research will aim to corroborate these results by extending this setting to more diverse datasets, such as medical images. In these applications, accurate probability estimation is crucial for segmentation, and the general scarcity of high-quality data poses a major challenge. Exploring the effectiveness of the CaPE method or other calibration techniques, such as semantic-aware grouping~\cite{yang2024beyond}, will provide deeper insights into its potential and limitations in these scenarios. This will ultimately contribute to developing more reliable and well-calibrated segmentation models.



\bibliographystyle{IEEEtran}
\bibliography{ref}

\end{document}